\documentclass[10pt,twocolumn,letterpaper]{article}

\usepackage{cvpr}
\usepackage{times}
\usepackage{epsfig}
\usepackage{graphicx}
\usepackage{amsmath}
\usepackage{amssymb}
\usepackage{url}
\usepackage{xcolor}
\usepackage{framed,multirow}

\DeclareMathOperator*{\argmin}{arg\,min}

\definecolor{newcolor}{rgb}{.8,.349,.1}

\usepackage[breaklinks=true,bookmarks=false]{hyperref}

\cvprfinalcopy %

\def\cvprPaperID{****} %

\begin{document}

\title{Rethinking Convolutional Features in Correlation Filter Based Tracking}

\author{
Fang Liang$^{1}$
\and
Wenjun Peng$^{1}$
\and
Qinghao Liu$^{2}$
\and
Haijin Wang$^{1}$
\and \\
$^{1}$Jiangxi Normal University, Nanchang, China
\and \\
$^{2}$The University of Adelaide, Adelaide, Australia
}

\maketitle

\begin{abstract}
Both accuracy and efficiency are of significant importance to the task of visual object tracking. In recent years, as the surge of deep learning, Deep Convolutional Neural Network (DCNN) becomes a very popular choice among the tracking community. However, due to the high computational complexity, end-to-end visual object trackers can hardly achieve an acceptable inference time and therefore can difficult to be utilized in many real-world applications. In this paper, we revisit a hierarchical deep feature-based visual tracker and found that both the performance and efficiency of the deep tracker are limited by the poor feature quality. Therefore, we propose a feature selection module to select more discriminative features for the trackers. After removing redundant features, our proposed tracker achieves significant improvements in both performance and efficiency. Finally, comparisons with state-of-the-art trackers are provided.
\end{abstract}

\section{Introduction}

Visual object tracking is one of the long standing computer vision tasks. The aim of visual tracking task is to estimate the trajectory of a target in a video sequence. Ground-truth bounding box of the target object is only given at the first frame of the video, then the tracker is asked to predict the target states (usually are position and size) at the following frames without any extra supervisions. Tracking of objects or keypoints has been widely used in robotics, video surveillance and intelligent traffic control system.

Benefit from Deep Convolutional Neural Networks (DCNNs), more and more deep learning based tracking approaches have been proposed in the last decade. Despite the well-admitted success, a dilemma still existing in the community is that, deep learning increases the tracking accuracy, while at the cost of high computational complexity. As a result, most top-performing deep trackers usually suffer from extremely low efficiency, which runs around 1 fps or even lower on high performance GPU accelerators \cite{danelljan2017eco, fan2017sanet, nam2016learning, pu2018deep, sun2018learning}. Recently, some real-time deep trackers are proposed \cite{bertinetto2016fully, he2018twofold, held2016learning, li2018high, zhang2018structured}. They achieve very fast tracking speed (more than 50 fps or even faster on GPU), however, a performance gap between these real-time trackers and top-performing trackers still exist.

In recent years, Discriminative Correlation Filter (DCF) based visual trackers have shown impressive improvements in terms of accuracy and robustness on public benchmarks \cite{wu2013online, wu2015object}. DCF-based trackers train a correlation filter to predict the target classification scores. Due to the high computational efficiency with the use of fast Fourier transforms, DCF-based trackers can efficiently utilize a number of circular-shifted versions of the training samples, which enables DCF-based methods to achieve competitive tracking performance with high inference speed. In conventional DCF-based methods, manual-crafted features such as HOG \cite{dalal2005histograms} and color attributes \cite{danelljan2014adaptive} are widely used for learning the correlation filters. As DCNNs show outstanding performance for many vision tasks in recent years, and is therefore of interest for DCF-based trackers. To get the benefit from such technologies, many DCF-based trackers \cite{ma2015hierarchical, wang2017robust, wang2017deep} started to employ DCNNs as a feature extractor for learning correlation filters. Features extracted by DCNNs preserve both spatial details and semantics, which are much discriminative than those hand-crafted features.

Some previous works \cite{li2013survey, wang2015understanding} have revealed the power of appearance model, and show that the appearance model (feature extractor) plays the most important role in visual tracking systems. In most DCNN-DCF based works \cite{danelljan2017eco, danelljan2016beyond, ma2015hierarchical}, features employed for training correlation filters are directly extracted by a deep neural network pretrained on image classification dataset without any fine-tuning. However, different from the image classification and object detection task, category information is less important in visual object tracking task. Thus considerable redundancies are included in these pre-trained models, which might be harmful to both tracking efficiency and accuracy. 

In \cite{wang2017robust}, Wang \emph{et al.} proposed a domain adaptation method to transfer the deep features from classification domain to tracking domain, where the individual objects, rather than the image categories. After removing the redundant information, their proposed approach achieved almost three times faster than the baseline tracker. However, although the authors claim that the domain adaptation can guide the DCNN to learn more useful representations for tracking task, their tracking performance slightly dropped compared to the baseline tracker. This may because features of high quality are simultaneously abandoned with redundancies at the domain transferring stage. Similar to \cite{wang2017robust}, Choi \emph{et al.} \cite{choi2018context} proposed an encoder-decoder structure to compress the deep features for high speed visual tracking. Their proposed multi-expert auto-encoders robustly compress raw deep convolutional features, which are then employed to learn a DCF-based tracker later. Different from the strategy used in \cite{danelljan2017eco, danelljan2016beyond, ma2015hierarchical}, which directly utilized the raw convolutional features learned from a classification task. \cite{choi2018context, wang2017robust} fine-tuned the DCNNs on new datasets, which effectively adapt the original image classification domain to the new object tracking domain.

In this paper, we revisit the DCNN-DCF framework, and propose a novel data-driven method to measure feature quality for visual tracking task. By utilizing this method, redundant channel features are pruned, which simultaneously improve the tracking accuracy and speed. The main contributions of this paper are as followings:

\begin{itemize}
    \item We analyzed widely used deep representations in the DCNN-DCF based object trackers, and devised a feature quality measurement function to estimate the feature quality in a numerical way. By employing the proposed quality function, we found that most deep features extracted by the ImageNet pretrained DCNN are less discriminative for visual tracking task. 
    \item To enhance the feature quality, we further propose a data-driven feature selection method to remove redundant information. Benefit from this technology, a large number of redundancies are dropped, which simultaneously improve the tracking accuracy and speed.
    \item Finally, we evaluate the proposed tracker on public benchmark, the experiment results show our proposed tracker achieved comparable performance to the state-of-the-art approaches.
\end{itemize}

\section{Related Work}

\subsection{Convolutional Features}

In recent years, convolution neural networks achieved surprisingly success thanks to their ability in automatic feature extraction. After training on a large image classification dataset \cite{deng2009imagenet}, DCNN gain ability to extract highly discriminative features, and obtained much better performance than manual-crafted features (\emph{e.g.} HOG \cite{dalal2005histograms} and SIFT \cite{Lowe_sift}) in a lot of vision applications, such as classifying objects into 1000 categories and object detection \cite{girshick2014rcnn, wang2019human}.

A typical convolutional neural network, \emph{e.g.}VGG-net \cite{simonyan2014very}, is consist of convolutional layers, activation layers, pooling layers and fully connected layers. 
In DCNN-DCF tracking systems, as the DCF-based trackers do not rely on the regression results outputted by the neural network, the fully connected layers can therefore be abandoned to accelerate inference speed. Moreover, features from pooling layers are hardly used in deep trackers because of the lack of spatial information. Another noteworthy point is, due to the design of DCNN structure, features from deeper layers carry more semantics while the shallower layers include more spatial information. In \cite{ma2015hierarchical}, the authors claim that the features carried with more semantics can be exploited to handle large appearance changes, while the spatial information can be used for avoiding target drifting. Therefore, using ensemble of features extracted by different DCNN levels permits the correlation filters to learn a more robust and accurate tracker.

\subsection{Correlation Filter Based Tracking}

Discriminative correlation filters have attracted considerable attention in object tracking community, because of their efficient learning process. DCF-based trackers regress all the circular-shifted training samples of the input features to a target Gaussian function, therefore, no hard threshold are needed for the target appearance. \cite{bolme2010visual} is a pioneer work, which learned a Minimum Output Sum of Squared Error (MOSSE) for fast visual object tracking. Then \cite{henriques2012exploiting} introduced an extension work based on the MOSSE, which is termed CSK, to exploit Circular Structure of tracking-by-detection with Kernels. Henriques \emph{et al.} \cite{henriques2015high} extended the CSK to a Kernelized Correlation Filters (KCF), which significantly improved the tracking speed. In some more recent works, appearance model attracts more attention. \cite{danelljan2014adaptive} proposed to use adaptive color-attributes to replace the conventional color representations for DCF-based trackers. \cite{ma2015hierarchical} firstly employed DCNN to extract hierarchical convolutional features for learning multiple correlation filters, which achieved impressive performance on public benchmarks.

\section{Methods}

In this section, we first give a brief introduction to correlation filter based trackers, then we analyze existing problems in DCNN-DCF based tracking systems and propose a novel method to measure feature quality for a single channel featuremap. By employing this method, we further drop redundant and less discriminative deep features at inference time, which significantly improve the tracking speed and accuracy.

\subsection{Correlation Filter}

Benefiting from the property of the circulant matrix in Fourier domain and fast Fourier transformation (FFT) technologies, correlation filters can be trained in a very short time, which leads to high tracking performance under low computational cost. Thus, typical correlation filter based trackers \cite{danelljan2014adaptive, li2019real, ma2015hierarchical, wang2017robust, wang2017deep} usually learn a binary classifier and estimate the translation of target object by searching the maximum value of correlation response map.

Compared to the conventional correlation filter based trackers, Ma \emph{et al.} \cite{ma2015hierarchical} proposed to exploit hierarchical convolutional features for learning deep visual tracker. Deep features extracted from different levels of DCNNs capture variety of information, \emph{i.e.}, the earlier layers encode more spatial details while the latter layers provide more semantics. \cite{ma2015hierarchical} exploits both spatial details and semantics from hierarchical convolutional features. 
In practical, the authors use a VGG-19 Net \cite{simonyan2014very} as feature extractor, and features extracted by three levels (\emph{Conv3-4}, \emph{Conv4-4} and \emph{Conv5-4}) are selected to train the correlation filters. Benefiting from the highly discriminative deep features and efficient kernelized correlation filters, \cite{ma2015hierarchical} achieved impressive tracking performance while keeping a practical speed.

Let $\textbf{x}$ denotes the feature vector extracted by the $l$-th layer from a DCNN backbone, which is a tensor of $M \times N \times D$ dimension, where $M$, $N$ and $D$ are width, height and channel numbers respectively. Then the circular shifted sample can be denoted as $\textbf{x}_{m, n}$, $(m, n) \in \{0, 1, ..., M - 1\} \times \{0, 1, ..., N - 1\}$. Each shifted sample has a Gaussian function label $y(m, n) = e^{-\frac{(m-M/2)^{2}+(n-N/2)^2}{2\sigma^{2}}}$, where $\sigma$ is the kernel width. The correlation filter $\textbf{w}$ then can be learned by solving the following minimization problem:

\begin{equation}
    \textbf{w}^{*} = \argmin_\textbf{w}\sum_{m, n}\|\textbf{w}\cdot\textbf{x}_{m, n}-y(m, n)\|^{2} + \lambda\|\textbf{w}\|^{2}_{2}
\label{eq:correlation_filter_target}
\end{equation}

\noindent where $\lambda (\lambda \geq 0)$ is a regularization parameter. Then the learned filter in the frequency domain on the $d$-th channel $l$-th layer writes:

\begin{equation}
    \textbf{W}^{d}_{l} = \frac{\textbf{Y} \odot \Bar{\textbf{X}}^{d}}{\sum_{i=1}^{D}\textbf{X}^{i} \odot \Bar{\textbf{X}}^{i} + \lambda}
\end{equation}

\noindent where $\textbf{Y}$ is the Fourier transformation form of $\textbf{y}$, and bar means complex conjugation. Given an image cropped by the searching window in next frame, which the convolutional feature of this image patch is denoted as $\textbf{z}$. Then the correlation response map of $l$-th layer can be written as:

\begin{equation}
    f_{l} = \mathcal{F}^{-1}(\sum_{d=1}^{D}\textbf{W}^{d} \odot \Bar{Z}^{d})
\end{equation}

\noindent where $\mathcal{F}^{-1}$ is inverse fast Fourier transformation. Then the location of the target object can be estimated by searching the position of the maximum value of the correlation response map.

\subsection{Convolutional Features}

\subsubsection{Zero Activation}

\begin{figure}[htb!]
    \centering
    \includegraphics[width=0.45\textwidth]{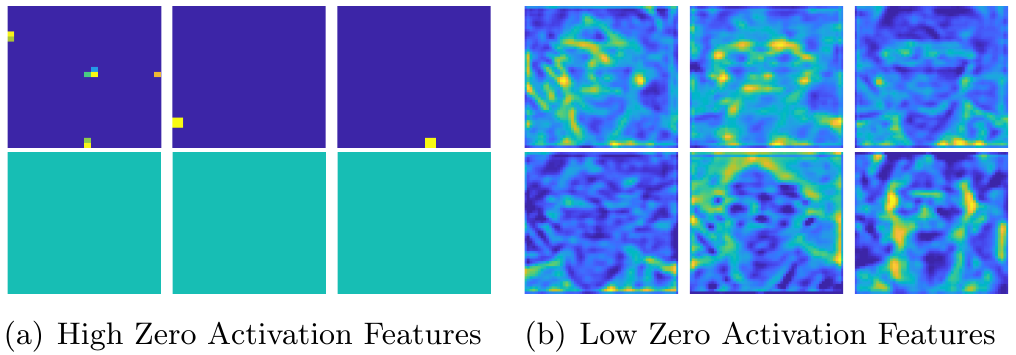}
    \caption{Features with different zero activation (better viewed in color).}
    \label{fig:features_comparison}
\end{figure}

During the last decade, a variety of deep neural network architectures are carefully designed by experienced experts. From AlexNet \cite{krizhevsky2012imagenet} to ResNet \cite{he2016deep}, state-of-the-art neural networks are getting deeper and wider. However, an interesting observation is that, outputs of a significant portion of neurons in a large deep convolutional neural network are mostly zero, regardless of what inputs the neural network received. Therefore, the featuremaps outputted by these zero activation neurons can be seen as redundancies, due to their poor discrimination (See Fig.~\ref{fig:features_comparison}). In \cite{hu2016network}, Hu \emph{et al.} proposed Average Percentage of Zeros (APoZ) for network trimming task. The authors claim that the neurons with zero activation can be removed without affecting the overall accuracy of the network. This so called APoZ is defined to measure the percentage of zero activations of a neuron after the activation function (\emph{e.g. ReLU}). Let $Feat_{c}^{(i)}$ denotes the output featuremap of $c$-th channel in $i$-th layer, then the $APoZ_{c}^{(i)}$ of the $c$-th neuron in $i$-th layer is defined as:

\begin{equation}
    APoZ_{c}^{(i)} = APoZ(Feat_{c}^{(i)}) = \frac{\sum_{k}^{N}\sum_{j}^{M}f(Feat_{c,j}^{(i)}(k) = 0)}{N \times M}
\label{eq:APOZ}
\end{equation}

\noindent where $f(\cdot) = 1$ if \emph{true}, and $f(\cdot) = 0$ if $false$, $M$ and $N$ are the dimension of featuremap of $Feat_{c}^{(i)}$ and the total number of validation examples respectively.

Different from the deep network trimming or channel pruning task, the classification performance of the network does not represent the final tracking performance. In DCNN-DCF based tracking approaches, convolutional neural networks are only utilized as feature extractor, where the fully connected layers of the networks are dropped, only convolution, activation and pooling layers are kept for extracting deep features. Thus the APoZ cannot be employed in tracking tasks directly.
Therefore, we define a percentage of zeros for single channel featuremap, which is termed SPoZ. Let $X_{c}^{(i)}$ denotes a single featuremap from the $c$-th channel $i$-th layer, which is a $W \times H$ matrix, where $W$ and $H$ are width and height respectively, each element from the $w$-th column and $h$-th row can be denoted as $X_{c}^{(i)}(w, h)$. Then the $SPoZ_{c}^{i}$ of the $c$-th channel in $i$-th layer writes:

\begin{equation}
SPoZ_{c}^{(i)} = SPoZ(X_{c}^{(i)}) = \frac{\sum_{w}^{W}\sum_{h}^{H}f(X_{c}^{(i)}(w, h) = 0)}{W \times H}
\label{eq:SPOZ}
\end{equation}

\noindent where $f(\cdot) = 1$ if \emph{true}, and $f(\cdot) = 0$ if \emph{false}. Specifically, when $SPoZ(X_{c}^{(i)}) = 1$, then $X_{c}^{(i)}$ is a zero activation featuremap which defined in the APoZ.

\begin{figure}[htb!]
    \centering
    \includegraphics[width=0.5\textwidth]{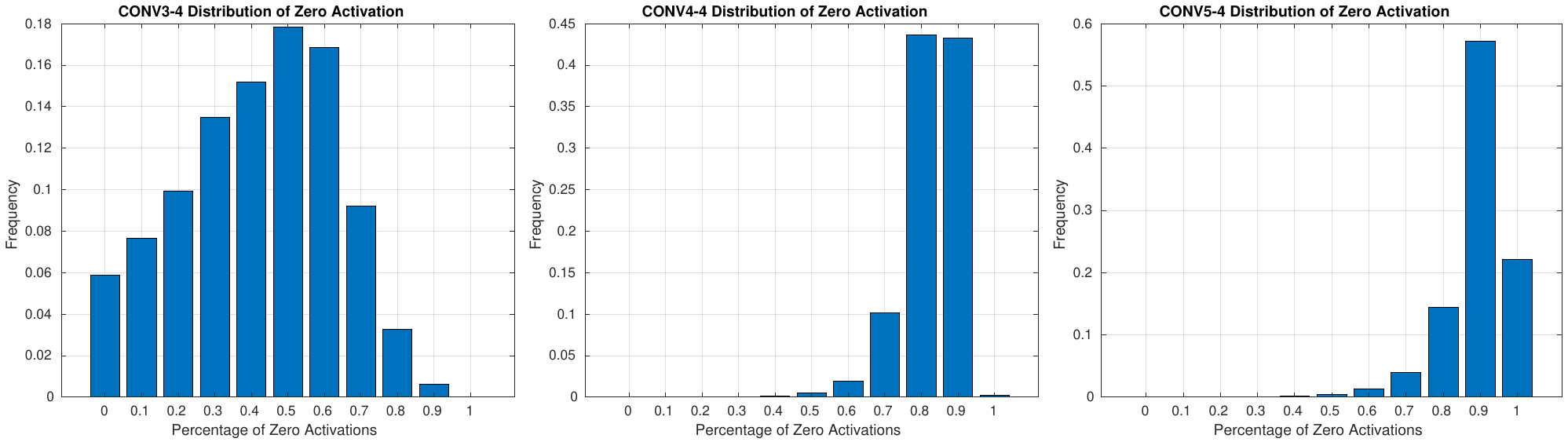}
    \caption{Distribution of SPoZ of three feature extracting layers (\emph{Conv3-4}, \emph{Conv4-4} and \emph{Conv5-4}) used in \cite{ma2015hierarchical}. The backbone network is VGG-19 Net \cite{simonyan2014very} trained on ImageNet dataset \cite{deng2009imagenet} with category-level label.}
    \label{fig:HCF_feat_zero_activation}
\end{figure}

Fig.~\ref{fig:HCF_feat_zero_activation} shows the distribution of SPoZ of the deep convolutional features extracted by three layers\footnote{The backbone network is VGG-19 and identical to the one used in \cite{ma2015hierarchical}} (\emph{Conv3-4}, \emph{Conv4-4} and \emph{Conv5-4}). As shown in Fig.~\ref{fig:HCF_feat_zero_activation}, features extracted by the deeper layers are consist of a large number of zero activations, which might include lots of redundant information.

\begin{table}[htb!]
\begin{center}
\begin{tabular}{cccccc}
\hline
Conv3-4     & Conv4-4       & Con5-4        & $C$ & $P$ & FPS \\ \hline\hline
\checkmark  &               &               & 256            & 0.756     & 35.7        \\ \hline
            & \checkmark    &               & 512            & 0.812     & 23.1        \\ \hline
            &               & \checkmark    & 512            & 0.849     & 21.3        \\ \hline
\checkmark  & \checkmark    & \checkmark    & 1280           & 0.891     & 11.0        \\ \hline
\end{tabular}
\end{center}
\caption{Tracking performance comparison between features extracted by single layer \emph{Conv3-4}, \emph{Conv4-4}, \emph{Conv5-4} and the ensemble features on CVPR2013 dataset \cite{wu2013online}. $C$ is channel number; $P$ indicates the average distance precision rate when the location error threshold equals 20 pixels (The higher the better); $S$ is running Speed.}
\label{table:single_layer_performance}
\end{table}

\subsubsection{Existing Problems}

As the features with high SPoZ may represent less discrimination, a natural idea is to use SPoZ as guidance to select convolutional features of high quality for training correlation filters. To explore the effectiveness, we implement a DCF based tracker which only employs single-layer convolutional features, and the comparison of tracking performance between \emph{Conv3-4}, \emph{Conv4-4} and \emph{Conv5-4} is given. As shown in Table~\ref{table:single_layer_performance}, tracking performance of each convolutional layer on CVPR2013 dataset \cite{wu2013online} is evaluated. However, the features from layer \emph{Conv3-4}, which have shown lower frequency of zero activations, do not promise a competitive tracking accuracy, while the features from layer \emph{Conv5-4}, which hold highest frequency of zero activation, achieve better performance. Therefore, the questions are: {\it Why the features extracted by latter convolutional layers show high zero activation on object tracking benchmark? And why the DCF-based tracker performs well by exploiting these features?}

We give possible answers as followings:

\begin{itemize}
    \item Data Distribution: Most feature extractors used in the DCNN-DCF based tracking systems are pretrained on the large ImageNet classification dataset, which contains more than 1 million images. Training on such a large dataset allows DCNN to learn highly discriminative features, each neuron in the network would be sensitive to different texture, edge, corner and object. However, the category of tracking objects is limited, while most tracking targets in real-world are moving objects such as human (pedestrian or face), vehicle and animal. Therefore, most neurons which sensitive to still objects, \emph{e.g.}, tree, building and furniture might be less possible to be activated, thus they have a higher chance to output zero activation. Moreover, another difference between the image classification and object tracking task is that, in most case, visual object tracker only search a small region which near to the target position predicted in the last frame, which means the current video frame would be cropped into a small image patch by a searching window, the target object is therefore almost located in the center of the image in most time, which led to a different object distribution compared to the classification dataset. Besides, in tracking tasks, the target object is not related to the object category, hence, a person can be a positive sample while another person can be a negative sample at the same video sequence, even though they belong to the same category and have similar appearance. All of these issues may introduce conflicts between classification and tracking tasks, which thus further led to a high frequency of zero activation.
    \item Effective Features: As it has been proved in some previous works \cite{choi2018context, wang2017robust}, a limited number of features can promise competitive tracking performance. Therefore, although features extracted by latter layers of the DCNN are consist of a large number of zero activations, highly discriminative features are still included, which contribute most to the final tracking performance. Meanwhile, the ensemble of multi-layer features allow the tracker to collect more useful information from a large feature pool, thus the tracker can be boosted by features from both shallower and deeper layers. However, even though the ensemble method provides more possibility for the trackers to extract more discriminative features, it introduces a large number of redundant and useless information at the same time, which may cause a slower running speed.
\end{itemize}

\begin{figure}[htb!]
    \centering
    \includegraphics[width=0.45\textwidth]{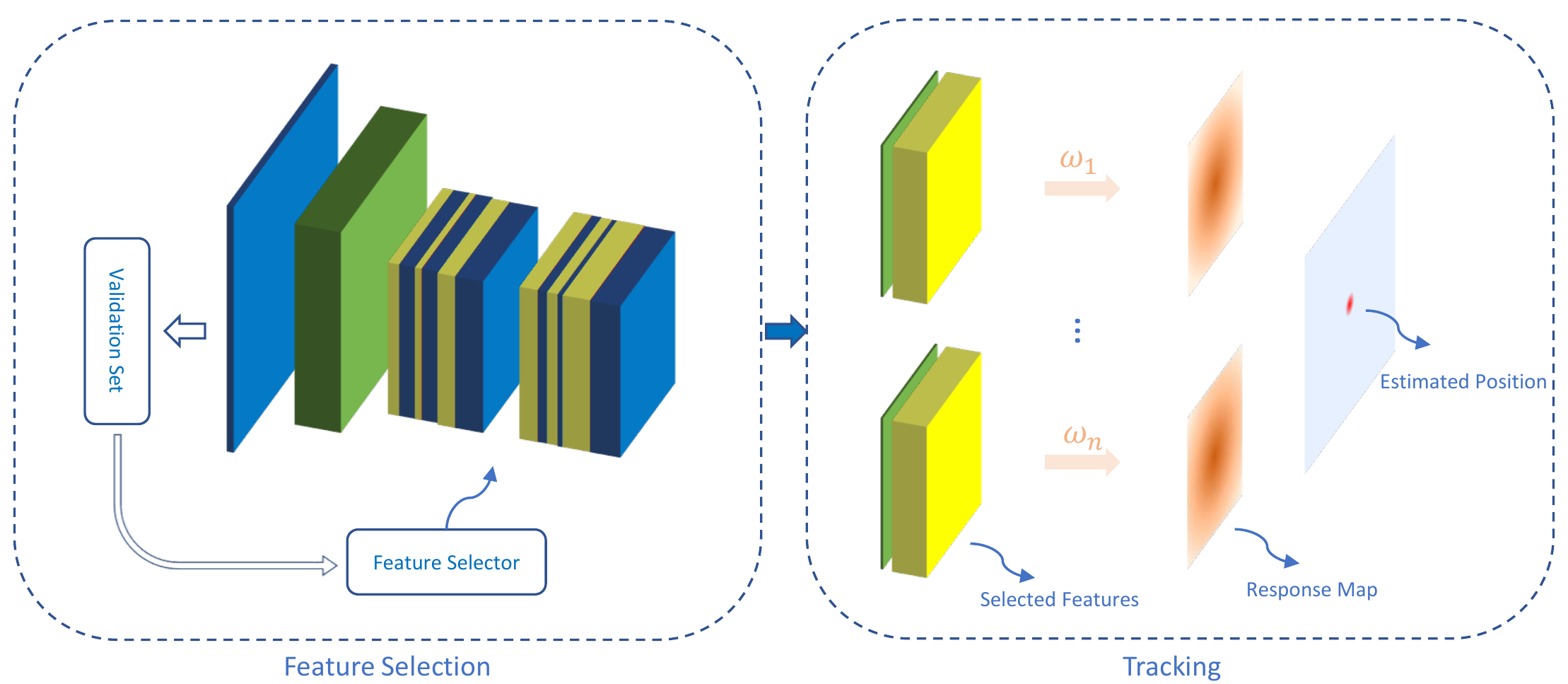}
    \caption{Structure of the proposed DAT tracker.}
    \label{fig:structure}
\end{figure}

To solve these problems, we propose a Discrimination-Aware Tracking (DAT) method (see Fig.~\ref{fig:structure}). An \emph{Feature Selection} module is installed to prune redundant channel features for the selected feature extraction layers.

\subsection{Feature Selection Strategy}

As described in Eq.~\ref{eq:SPOZ}, we propose SPoZ to measure single channel feature discrimination. However, the experiment results in Table~\ref{table:single_layer_performance} show that simply employ this SPoZ to describe feature quality cannot be helpful for selecting deep features. After further analysis, we found that a large amount of featuremaps with high APoZ only activate to the background areas, which lacks appearance details of the target object. It is true background samples are also important to the object tracker, however, most foreground featuremaps have already carried a part of background information. Thus those channels which only encoded with background objects can be dropped. Therefore, to filter these redundant background features, we propose a foreground-aware measurement to calculate the ratio between foreground activation and background activation, which is termed Activation Ratio. Because the SPoZ can only used to measure zero activation, we first rewrite it to an energy function to calculate average area activation by remove the $f(\cdot)$ in Eq.~\ref{eq:SPOZ}:

\begin{equation}
e(X_{c}^{(i)}) = \frac{\sum_{w}^{W}\sum_{h}^{H}X_{c}^{(i)}(w, h)}{W \times H}
\label{eq:SPOZ_new}
\end{equation}

Let $W$ and $H$ denotes the width and height of the input image patch respectively, and $b_n = [x_n, y_n, w_n, h_n]$ indicates the ground-truth bounding box of the target object in $n$-th image in validation dataset\footnote{To avoid possible over-fitting, we use another widely used tracking dataset \cite{VOT_TPAMI} as validation set.}. Then the features from $i$-th layer $c$-th channel which include foreground area are denoted as $X_{b_{n}}^{(i), c}$, the background features from the same channel are denoted as $\Bar{X}_{b_{n}}^{(i), c}$. Thus the Activation Ratio of $i$-th layer $c$-th channel in $n$-th image can be written as:

\begin{equation}
r_{n}^{(i), c} = \frac{e(X_{b_{n}}^{(i), c})}{1 + e(\Bar{X}_{b_{n}}^{(i), c})} = \frac{\frac{1}{w_{n} \times h_{n}}\sum_{w}^{w_n}\sum_{h}^{h_n}x_{w, h}}{1 + \frac{1}{\bar{w}_{n} \times \Bar{h}_{n}}\sum_{w^{'}}^{\Bar{w}_{n}}\sum_{h^{'}}^{\Bar{h}_{n}} \Bar{x}_{w^{'},h^{'}}}
\label{eq:FPOZ}
\end{equation}

\noindent Where $x_{w,h} \in X_{b_{n}}^{(i), c}$ and $\Bar{x}_{w^{'},h{'}} \in \Bar{X}_{b_{n}}^{(i), c}$. Since we have defined the Activation Ratio for a single channel featuremap, then we can calculate an Average Activation Ratio for each channel on the validation set. The Average Activation Ratio of the $i$-th layer $c$-th channel writes:

\begin{equation}
R_{c}^{(i)} = \frac{\sum_{n}^{N}r_{n}^{(i), c}}{N}
\label{eq:AAR}
\end{equation}

\noindent At inference time, the features from a low Average Activation Ratio channel ($R_{c}^{(i)} < \eta_{i}$) would be dropped to enhance the feature quality.

\begin{figure}[htb!]
    \centering
    \includegraphics[width=0.45\textwidth]{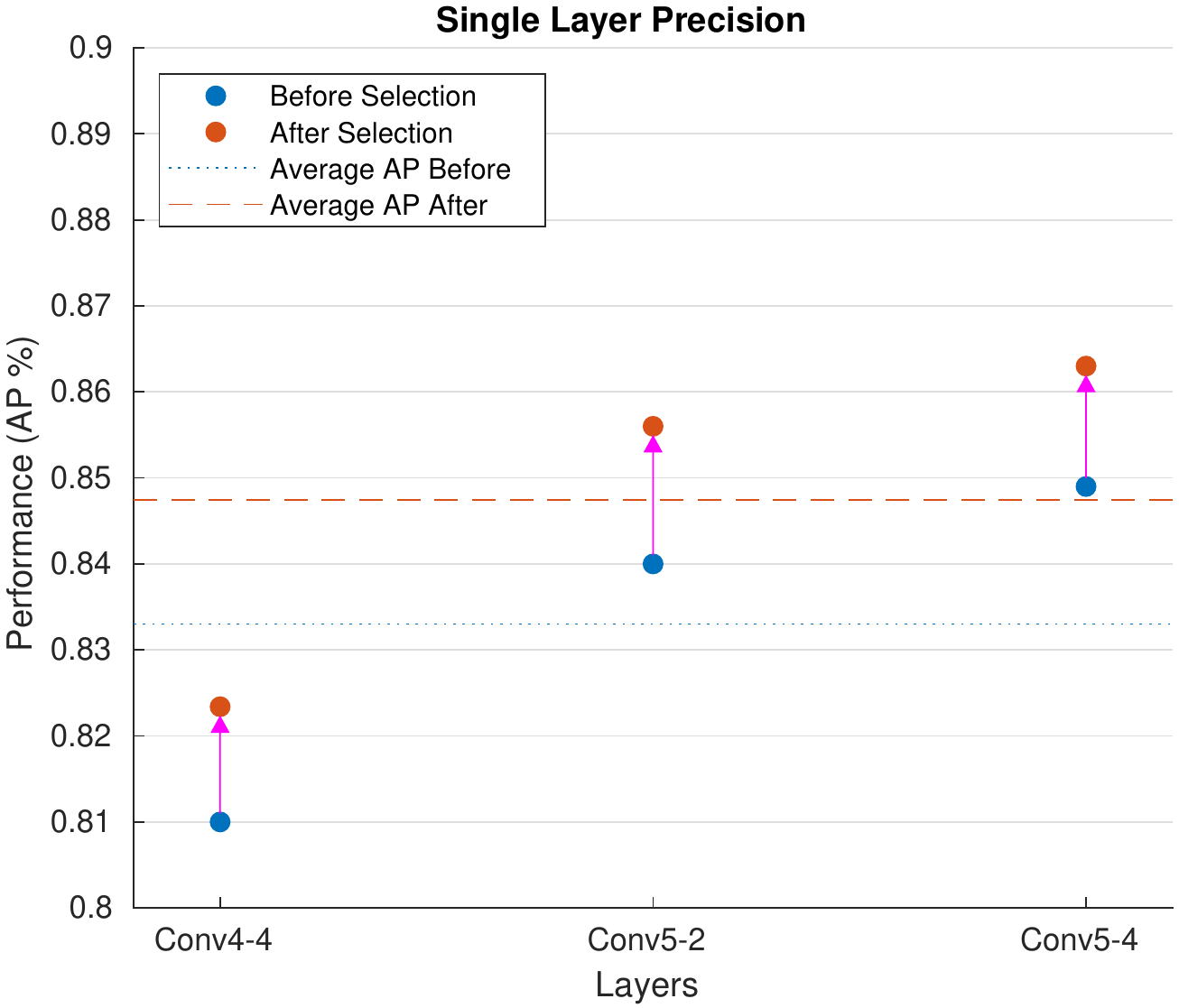}
    \caption{Single-layer convolutional feature based tracking performance on CVPR2013 dataset \cite{wu2013online}. \textcolor[rgb]{0,0.4431,0.7373}{Blue Points}: Tracking AP on CVPR2013 dataset before feature selection. \textcolor[rgb]{0.8471,0.3216,0.0941}{Red Points}: Tracking AP on CVPR2013 dataset after feature selection. \textcolor[rgb]{0,0.4431,0.7373}{Blue Dash Line}: Average AP of the three layers on the dataset before feature selection. \textcolor[rgb]{0.8471,0.3216,0.0941}{Red Dash Line}: Average AP of the three layers on the dataset after feature selection.}
    \label{fig:p_comparison}
\end{figure}

To evaluate the effectiveness of the proposed feature selection strategy, we compare single-layer convolutional feature based tracking performance on the CVPR2013 benchmark \cite{wu2013online}. The DCNN model used to extract deep convolutional features is identical to the one which was employed in \cite{ma2015hierarchical}. We evaluate tracking performance from three different layers, \emph{i.e.}, \emph{Conv4-4}, \emph{Conv5-2} and \emph{Conv5-4}. The \emph{Conv3-4} layer employed in \cite{ma2015hierarchical} did not make any contribution to the final performance, thus we use \emph{Conv5-2} to replace it here. As shown in Fig.~\ref{fig:p_comparison}, after adopting the feature selection strategy, tracking performance of all the three layers gain around 1.5\% improvements. The proposed feature selection strategy significantly improves the average tracking performance of these three layers from about 0.832 to 0.848. However, it is noteworthy that, the final tracking performance relies on not only one layer feature, thus the responses from each layer can make contributions to the final response map. Consequently, the final features from different layers may carry similar information, which is also redundant. To avoid this problem, we set larger threshold $\eta_{i}$ to shallower layers in the ensemble version, thus more channel features would be pruned. This strategy induces to a significant reduction of single-layer tracking performance for the shallower layers, however the overall tracking accuracy was not affected a lot.

\section{Experiments}

In this section, we show and compare our proposed tracking method DAT with the state-of-the-art trackers on two widely used tracking benchmarks \cite{wu2013online, wu2015object}. Both quantitative results and qualitative results are provided.

\subsection{Dataset and Evaluation Metrics}

We evaluate our proposed method on two widely used benchmarks, \emph{i.e.}, CVPR2013 \cite{wu2013online} (includes 51 targets in 50 videos), which contain the ground-truth bounding box of target object at every frame. These datasets are the most frequently used benchmarks among the visual object tracking community.

For performance measure, we follow the widely used Average Precision (AP) curve of one-pass evaluation (OPE) which proposed in \cite{wu2013online}. The AP curve was estimated by averaging the precisions of all video sequences, which includes two sources: distance precision rate (location error) and success rate (overlap). For convenient comparison with other state-of-the-art trackers, the average precisions when the location error threshold equals 20 pixels and the Area Under the Curve (AUC) of the success rate are used.

\subsection{Quantitative Results}

\begin{table}[htb!]
\begin{center}
\begin{tabular}{|c||c|c|c|c|}
\hline
                          & Algorithm                        & Precision & Speed (FPS) & GPU \\ \hline\hline
\multirow{1}{*}{} 
                          & DAT (Ours)                       & \textbf{0.910} & \textbf{38.5}  & Y \\ \cline{2-5}\hline\hline
\multirow{12}{*}{\rotatebox{90}{DCF-Based}} 
                          & MCPF~\cite{Zhang2017Multi}       & \textbf{0.916} & 0.6   & Y \\ \cline{2-5}
                          & HCF~\cite{ma2015hierarchical}    & 0.891 & 11.0  & Y \\ \cline{2-5}
                          & HDT~\cite{qi2016hedged}          & 0.889 & 10.0  & Y \\ \cline{2-5}
                          & SACF(D)~\cite{Zhang2018ECCV}     & 0.886 & 23.0  & Y \\ \cline{2-5} 
                          & MSDAT~\cite{wang2017robust}      & 0.881 & 34.8  & Y \\ \cline{2-5} 
                          & ACFN~\cite{Choi2017Attentional}  & 0.860 & 15.0  & Y \\ \cline{2-5}
                          & MCCTH~\cite{NingCVPR2018}        & 0.856 & 44.8  & Y  \\ \cline{2-5}
                          & DeepSRDCF~\cite{Danel2016Conv}   & 0.849 & 0.2   & N  \\ \cline{2-5} 
                          & MKCFup~\cite{Tang2018High}       & 0.835 & 150.0 & N \\ \cline{2-5} 
                          & CACF~\cite{Mueller2017Context}   & 0.833 & 35.2  & N \\ \cline{2-5} 
                          & CSR-DCF~\cite{Lukezic2017Dis}    & 0.800 & 13.0  & N \\ \cline{2-5}
                          & KCF~\cite{henriques2015high}     & 0.741 & \textbf{245.0} & N \\ \hline\hline 
\multirow{7}{*}{\rotatebox{90}{End-to-End}} 
                          & SANet~\cite{fan2017sanet}        & \textbf{0.950} & 1.0   & Y \\ \cline{2-5}
                          & MD-Net~\cite{nam2016learning}    & 0.948 & 1.0   & Y \\ \cline{2-5}
                          & DNT~\cite{Chi2017Dual}           & 0.907 & 5.0   & Y \\ \cline{2-5} 
                          & ADNet~\cite{Yun2017Action}       & 0.903 & 2.9   & Y \\ \cline{2-5} 
                          & DeepTrack~\cite{Li2016DeepTrack} & 0.826 & 3.0   & Y \\ \cline{2-5} 
                          & SiamFC~\cite{bertinetto2016fully}& 0.815 & 58.0  & Y \\ \cline{2-5} 
                          & GOTURN~\cite{held2016learning}   & 0.625 & \textbf{165.0} & Y \\ \hline
\end{tabular}    
\end{center}
\caption{\newline Quantative results on the CVPR2013 dataset \cite{wu2013online}.}
\label{table:cvpr13_results}
\end{table}

Table~\ref{table:cvpr13_results} shows the comparison of tracking performance between our proposed DAT tracker with the state-of-the-art correlation filter based trackers: MCPF~\cite{Zhang2017Multi}, HCF~\cite{ma2015hierarchical}, SACF~\cite{Zhang2018ECCV}, MSDAT~\cite{wang2017robust}, ACFN~\cite{Choi2017Attentional}, MCCTH~\cite{NingCVPR2018}, DeepSRDCF~\cite{Danel2016Conv}, MKCFup~\cite{Tang2018High}, CACF~\cite{Mueller2017Context}, CSR-DCF~\cite{Lukezic2017Dis}, KCF~\cite{henriques2015high}; and End-to-End deep learning based trackers: SANet~\cite{fan2017sanet}, MD-Net~\cite{nam2016learning}, DNT~\cite{Chi2017Dual}, ADNet~\cite{Yun2017Action}, DeepTrack~\cite{Li2016DeepTrack}, SiamFC~\cite{bertinetto2016fully}, GOTURN~\cite{held2016learning} on CVPR2013~\cite{wu2013online} benchmark.

As shown in the results, top performing trackers of both DCF-based and End-to-End based methods, \emph{i.e.}, MCPF, SANet and MD-Net suffer from extremely slow inference speed. These three trackers achieve impressive tracking accuracy on the CVPR2013 dataset, but can hardly be used in real world applications. On the contrary, GOTURN trained DCNN on a very large video dataset, and abandoned all the online updating process, thus achieved very fast tracking speed. KCF learn correlation filters from hand-crafted features, which is much more efficient than deep learning technology. However, most high speed trackers such as GOTURN, SiamFC, KCF and MKCFup cannot obtain competitive results on the public benchmarks. Our proposed DAT tracker achieves similar performance to the MCPF tracker on the CVPR2013 benchmark, but much faster.

\section{Conclusion}
 
In this paper, we diagnosed and analyzed the DCNN-DCF based tracking system, and found that the deep features extracted by image classification task pretrained model are consist of a large number of redundant and harmful information for the tracking task. Inspired by network channel pruning task, we found that features from deeper layers of the DCNN show high percentage of zero activation on the tracking dataset, these zero activation features are less discriminative and can make a limited contribution to the tracking accuracy. Therefore, we propose a feature quality measuring method for single channel features. By employing this method, we use a validation set to calculate average activation ratio for each channel in the DCNN feature extractor. Then we propose a discrimination-aware tracking method, which is termed DAT. After removing those redundant features, our DAT tracker achieves significant improvements on tracking benchmark. However, the proposed DAT algorithm requires to pre-compute average activation ratio on a large validation set, which consumes a large number of computational resources. Thus, one of the possible future direction is to design an online feature selecting strategy, which can automatically and adaptively select channel features for different target objects and video sequences at inference time.

\end{document}